\documentclass[twoside]{article}

\usepackage[accepted]{aistats2020}
\usepackage{graphicx}
\usepackage{amsmath,amsthm,amssymb}
\usepackage{algorithm}
\usepackage[noend]{algpseudocode}
\usepackage{color}
\usepackage{scrextend}
\usepackage[table,xcdraw]{xcolor}
\usepackage{multirow}
\usepackage{enumerate}  
\usepackage{enumitem}
\usepackage{subcaption}
\usepackage{url}
\usepackage{natbib}

\DeclareMathOperator*{\argmax}{arg\,max}
\DeclareMathOperator*{\argmin}{arg\,min}

\renewcommand{\P}{\mathbb{P}}
\newcommand{\E}{\mathbb{E}}

\newcommand{\R}{\mathbb{R}}

\newcommand{\Nc}{\mathcal{N}}
\newcommand{\Cc}{\mathcal{C}}

\newcommand{\Fc}{\mathcal{F}}
\newcommand{\Dc}{\mathcal{D}}
\newcommand{\Xc}{\mathcal{X}}
\newcommand{\Oc}{\mathcal{O}}
\newcommand{\Pc}{\mathcal{P}}
\newcommand{\Qc}{\mathcal{Q}}
\newcommand{\Uc}{\mathcal{U}}
\newcommand{\Hc}{\mathcal{H}}

\newcommand{\T}{\top}

\newcommand{\yb}{\mathbf{y}}

\newcommand{\ucb}{\mathrm{ucb}}
\newcommand{\lcb}{\mathrm{lcb}}

\newcommand{\<}{\langle}
\renewcommand{\>}{\rangle}

\newtheorem{theorem}{Theorem}

\newtheorem{lemma}[theorem]{Lemma}
\newtheorem{corollary}[theorem]{Corollary}

\usepackage{printlen}

\begin{document}



\twocolumn[

\aistatstitle{Distributionally Robust Bayesian Optimization}

\aistatsauthor{ Johannes Kirschner \And Ilija Bogunovic \And  Stefanie Jegelka \And Andreas Krause }
\aistatsaddress{ ETH Z\"urich \And  ETH Z\"urich \And MIT \And ETH Z\"urich } ]

%\aistatsauthor{Anonymous Authors}
%\aistatsaddress{Anonymous Institution} 

\begin{abstract}
Robustness to distributional shift is one of the key challenges of contemporary machine learning. Attaining such robustness is the goal of distributionally robust optimization, which seeks a solution to an optimization problem that is worst-case robust under a specified distributional shift of an uncontrolled covariate. In this paper, we study such a problem when the distributional shift is measured via the maximum mean discrepancy (MMD). For the setting of zeroth-order, noisy optimization, we present a novel distributionally robust Bayesian optimization algorithm (DRBO). Our algorithm provably obtains sub-linear robust regret in various settings that differ in how the uncertain covariate is observed. We demonstrate the robust performance of our method on both synthetic and real-world benchmarks.\looseness=-1
\end{abstract}

\section{Introduction}

Bayesian optimization (BO) is a framework for model-based sequential optimization of \emph{black-box} functions that are expensive to evaluate and for which noisy point evaluations are available. Bayesian optimization algorithms have been successfully applied in a wide range of applications where the goal is to discover best-performing designs from a small number of trials, e.g., in vaccine and molecular design, gene optimization, automatic machine learning, robotics and control tasks, and many more.

\begin{table}[t]
	\centering
	\begin{tabular}{|c|c|}
		\hline
		\textbf{Objective} & \textbf{Formulation} %& \textbf{Algorithm / Examples ?} 
		\\ \hline
		\cellcolor[HTML]{EFEFEF} &  %&  
		\\
		\cellcolor[HTML]{EFEFEF} &  %&  
		\\
		\multirow{-3}{*}{\cellcolor[HTML]{EFEFEF}\begin{tabular}[c]{@{}c@{}}\textbf{Stochastic (SO)}\end{tabular}} & \multirow{-3}{*}{$\max\limits_x \mathbb{E}_{c\sim P}	[f(x,c)]$} \multirow{-3}{*}{} %&  \multirow{-3}{*}{\begin{tabular}[c]{@{}c@{}}Vaccine design~\cite{krause2011contextual}, \\ Crop recommendation~\cite{kirschner2019stochastic}\end{tabular}} 
		\\ \hline
		\cellcolor[HTML]{EFEFEF} & \cellcolor[HTML]{FFFFFF} %& \cellcolor[HTML]{FFFFFF} 
		\\
		\cellcolor[HTML]{EFEFEF} & \cellcolor[HTML]{FFFFFF} %& \cellcolor[HTML]{FFFFFF} 
		\\
		\multirow{-3}{*}{\cellcolor[HTML]{EFEFEF}\begin{tabular}[c]{@{}c@{}}\textbf{Worst-case }\\ \textbf{robust (RO)}\end{tabular}} &  \multirow{-3}{*}{\cellcolor[HTML]{FFFFFF} $\max\limits_x \min\limits_{c \in \Delta} f(x,c)$} \multirow{-3}{*}{\cellcolor[HTML]{FFFFFF}} %& \multirow{-3}{*}{\cellcolor[HTML]{FFFFFF}\begin{tabular}[c]{@{}c@{}}Lake monitoring and  \\ movie recommendation~\cite{bogunovic2018adversarially}\end{tabular}} 
		\\ \hline
		\cellcolor[HTML]{C0C0C0} & \cellcolor[HTML]{FFFFFF} %& \cellcolor[HTML]{FFFFFF}
		\\
		\cellcolor[HTML]{C0C0C0} & \cellcolor[HTML]{FFFFFF} %& \cellcolor[HTML]{FFFFFF} 
		\\
		\multirow{-3}{*}{\cellcolor[HTML]{C0C0C0}\begin{tabular}[c]{@{}c@{}}\textbf{Distributionally}\\{\textbf{robust (DRO)}}\end{tabular}} &  \multirow{-3}{*}{\cellcolor[HTML]{FFFFFF} $\max\limits_x \inf\limits_{\Qc \in \Uc} \E_{c \sim Q} [f(x,c)]$} \multirow{-3}{*}{\cellcolor[HTML]{FFFFFF}} %& \multirow{-3}{*}{\cellcolor[HTML]{FFFFFF}\begin{tabular}[c]{@{}c@{}}Lake monitoring and  \\ movie recommendation~\cite{bogunovic2018adversarially}\end{tabular}} 
		\\ \hline
	\end{tabular}%
	\vspace{-5pt}
	\caption{Different optimization objectives considered in Bayesian optimization.}
	\label{table:objectives}
\end{table}

In many practical tasks, the objective also depends on \emph{contextual} covariates of the environment. If this context follows a known distribution, the setting is essentially that of stochastic optimization with the objective to maximize the expected pay-off. Often, however, there exists a distributional mismatch between the covariate distribution that the learner assumes, and the true distribution of the environment. Examples include automated machine learning, where hyperparameters are tuned on training data while the test distribution can differ; recommender systems, where the distribution of the users shifts with time; and robotics, where the simulated environmental variables are only an approximation of the real physical world. In particular, whenever there is a distributional mismatch between the true and the data distribution used at training time, the optimization solutions can result in inferior performance or even lead to unsafe/unreliable execution. The problem of \emph{distributional data shift} has been recently identified as one of the most prevalent concrete challenges of modern AI safety~\citep{amodei2016concrete}. While the connection of robust optimization (RO) and Bayesian optimization has recently been established by \citet{bogunovic2018adversarially}, robustness to \emph{distributional data shift} remains unexplored in this field.

In this paper, we introduce the setting of {\em distributionally robust Bayesian optimization (DRBO)}: The goal is to track the optimal input that maximizes the expected function value under the worst-case distribution of an external, contextual parameter. In distributionally robust optimization (DRO), such a worst-case distribution belongs to a known \emph{uncertainty set} of distributions that is typically chosen as a ball centered around a given reference distribution. To measure the distance between distributions, in this work, we focus on the kernel-based \emph{maximum mean discrepancy} (MMD) distance. This metric fits well with the kernel-based regularity assumptions on the unknown function that are typically made in Bayesian optimization. 
%In this setting, we present novel robust \emph{no-regret} learning algorithms that differ in how the context parameter is observed.          

\subsection{Related Work}
\looseness=-1 A large number of Bayesian optimization algorithms have been developed over the years,  \citep[e.g.][]{srinivas2009gaussian,wang2017max,hennig2012entropy,chowdhury17kernelized,bogunovic2016truncated}. Several practical variants of the standard setting were addressed recently, including contextual~\citep{krause2011contextual,valko2013finite,lamprier2018profile,kirschner2019context} and time-varying~\citep{bogunovic2016time} BO, high-dimensional BO~\citep{djolonga2013high,kandasamy2015high, kirschner2019adaptive}, BO with constraints~\citep{gardner2014bayesian,gelbart2014bayesian}, heteroscedastic noise~\citep{kirschner18heteroscedastic} and uncertain inputs~\citep{oliveira2019bayesian}.    %Bayesian optimization \cite{srinivas2009gaussian}, \cite{abbasi2013online}, \cite{chowdhury17kernelized}  
% uncertain input \cite{lamprier2018profile}

Two classical objectives for optimization under uncertainty are stochastic optimization (SO) \citep{srinivas2009gaussian,krause2011contextual,lamprier2018profile,oliveira2019bayesian,kirschner2019context} and robust optimization (RO) \citep{bogunovic2018adversarially}, see Table~\ref{table:objectives}. SO asks for a solution that performs well in expectation over an uncontrolled, stochastic covariate. Here, the assumption is that the distribution of the contextual parameter is known, or (i.i.d.)\ samples are provided. Some variants of SO have been considered in the related contextual Bayesian optimization works~\citep{krause2011contextual, valko2013finite,kirschner2019context}. RO aims at a solution that is robust with respect to the worst possible realization of the context parameter. 
The RO objective has recently been studied in Bayesian optimization in~\citep{bogunovic2018adversarially}; the authors provide a robust BO algorithm, and obtain strong regret guarantees. In many practical scenarios, however, the solution to the SO problem might be highly \emph{non-robust}, while on the other hand, the worst-case RO solution might be overly \emph{pessimistic}. This motivates us to consider the \emph{distributionally robust optimization} (DRO), which is a ``middle ground'' between SO and RO. \looseness=-1

\looseness=-1 Distributionally robust optimization (DRO) dates back to the seminal work of~\citet{scarf1957min} and since then it has become an important topic in robust optimization \citep[e.g.][]{bertsimas2018data,goh2010distributionally}. It has recently received significant attention in machine learning, in particular due to its relation to regularization, adversarial learning, and generalization~\citep{staib2018distributionally}. The full literature on DRO is too vast to be adequately covered here, so we refer the interested reader to the recent review by \citet{rahimian2019distributionally} and references within.
For defining the uncertainty sets of distributions, different DRO works have studied $\phi$-divergences~\citep{ben2013robust,namkoong2017variance}, Wasserstein~\citep{gao2017wasserstein,esfahani2018data,sinha2017certifying} and the MMD~\citep{staib2019kernel} distances. 
In this work, we focus on the kernel-based MMD distance, but unlike previous DRO works, we assume that the objective function is \emph{unknown}, and only noisy point evaluations are available. 

We conclude this section by mentioning other robust aspects and settings that have been previously considered in Bayesian optimization. BO with outliers has been considered by \citet{martinez2017practical}, while the setting in which sampled points are subject to uncertainty has been studied by \citet{nogueira2016unscented,beland2017bayes,oliveira2019bayesian}. These settings differ significantly from the one considered in this paper and they do not consider robustness under distributional shift. Finally, we note that another robust BO algorithm has been recently developed for playing unknown repeated games against non-cooperative agents~\citep{sessa2019noregret}. 

While this work was under submission, a related approach for distributionally robust Bayesian quadrature appeared online \citep{nguyen2020distributionally}. The authors propose an approach based on Thompson sampling to solve a related robust objective for Bayesian quadrature. Our work captures this scenario in the "simulator setting", detailed below. The main difference in the analysis is that we bound worst-case frequentist regret opposed to the expected Bayesian regret.

\paragraph{Contributions} We propose a novel, distributionally robust Bayesian optimization (DRBO) algorithm.
	Our analysis shows that the DRBO achieves sublinear robust regret on several variants of the setting. Finally, we demonstrate robust performance of the DRBO method on synthetic and real-world benchmarks.

\section{Problem Statement}
Let $f: \Xc \times \Cc \rightarrow \R$ be an \emph{unknown} reward function defined over a parameter space $\Xc \times \Cc$ with finite\footnote{Our formulation and the theory extend to continuous sets $\Cc$ and $\Xc$, but for the algorithm we rely on solving a convex program of size $|\Cc|$.} action and context sets, $\Xc$ and $\Cc$. The objective is to optimize $f$ from \emph{sequential} and \emph{noisy} point evaluations. In our main setup, at each time step $t$, the learner chooses $x_t \in \Xc$ whereas the environment provides the context $c_t \in \Cc$ together with the \emph{noisy} function observation $y_t = f(x_t, c_t) + \xi_t$, where $\xi_t \sim \Nc(0, \sigma^2)$ with known $\sigma^2$ and independence between time steps. More generally, our results hold if the noise is $\sigma$-sub-Gaussian, which allows for non-Gaussian likelihoods (e.g.,~bounded noise). Further, we assume that $c_t$ is sampled independently from an unknown, time-dependent distribution $P_t^*$. 

\textbf{Optimization objective.} We consider the \emph{distributionally robust optimization} (DRO)~\citep{scarf1957min} objective, which asks to perform well simultaneously for a range of problems, each determined by a distribution in some uncertainty set. This is in contrast to SO, where we seek good performance against a single problem instance parametrized by a given distribution.

\begin{figure}[t]
	\centering
	\includegraphics{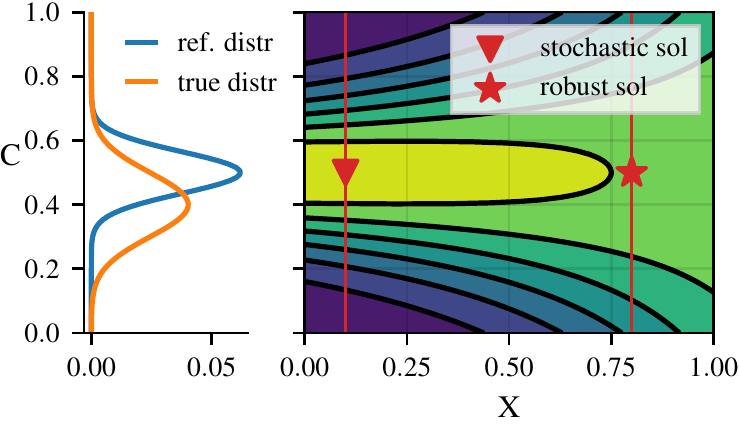}
	\caption{A function where the robust solution significantly differs from the stochastic solution. The learner obtains the blue reference distribution over the context set $\Cc$ and chooses a design $x \in \Xc$. If the distribution over the context set is equal to the reference, the solution marked by the triangle maximizes the expected reward. On the other hand, if the true distribution (orange) is shifted away from the reference, the flatter region of the reward function, marked by the star, provides higher expected reward.} \label{fig: benchmark 1 illustration}
\end{figure}

In DRO, the objective is to find $x \in \Xc$ that solves % the following problem:
\begin{equation} \label{eq:dro}
	\max_{x\in \Xc} \inf_{Q \in \Uc_t} \E_{c\sim Q}[f(x, c)].
\end{equation}
Here, $\Uc_t$ is a known \emph{uncertainty set} of distributions over $\Cc$ that can depend on the step $t$ and contains the true distribution $P_t^* \in \Uc_t$. Typically, $\Uc_t$ is chosen as a ball of radius (or margin) $\epsilon_t > 0$, and centered around a given \emph{reference distribution} $P_t$ on $\Cc$, i.e., $$\Uc_t = \lbrace Q: d(Q,P_t) \leq \epsilon_t \rbrace,$$
where $d(\cdot,\cdot)$ measures the discrepancy between two distributions. A possible choice for the reference distribution $P_t$, is the empirical sample distribution $\hat{P}_t = t^{-1} \sum_{s=1}^{t} \delta_{c_s}$, which is an instance of \emph{data-driven DRO} \citep{bertsimas2018data}. Depending on the underlying function and the uncertainty set $\Uc_t$, the robust solution can significantly differ from the solution to the stochastic objective $\max_{x \in \Xc} \E_{c \sim P}[f(x,c)]$ for a fixed (and typically known) distribution $P$. We illustrate such a case in Fig.\ \ref{fig: benchmark 1 illustration}.

Hence, at time step $t$, the learner receives a reference distribution $P_t \in \Pc(C)$ and margin $\epsilon_t > 0$. Our objective is to choose a sequence of actions $x_1,\dots,x_T$ that minimizes \emph{robust cumulative regret}:
\begin{equation}
	\label{eq:robust_cumul_regret}
	R_T = \sum_{t=1}^T \inf_{Q \in \Uc_t} \E_{Q}[f(x_t^{*},c)] - \inf_{Q \in \Uc_t} \E_{Q} [f(x_t,c)],
\end{equation}
where $x_t^{*} = \max_{x \in \Xc} \inf_{Q \in \Uc_t} \E_{Q}[f(x,c)]$. The \emph{robust regret} measures the cumulative loss of the learner on the chosen sequence of actions w.r.t.\ the worst case distribution over $\Cc$.

\paragraph{RKHS Regression.} The main regularity assumption of Bayesian optimization is that $f$ belongs to a reproducing kernel Hilbert space (RKHS) $\Hc$ with known kernel $k$. We denote the Hilbert norm by $\|\cdot\|_\Hc$  and assume $\| f \|_\Hc \leq B$ for some known $B > 0$.
From the observed data $\Dc_t = \{(x_1,c_1, y_1), \dots, (x_t,c_t,y_t)\}$, we can compute a kernel ridge regression estimate with
\begin{equation}
	\hat{f}_t = \argmin_{g \in \Hc} \sum_{i=1}^{t-1} (g(x_i,c_i) - y_i)^2 + \|g\|_{\Hc}^2 \text{ .} \label{eq: kernel ridge regression}
\end{equation}
The representer theorem provides the standard, closed-form solution for the least-squares estimate \citep{rasmussen2006gaussian}. % (Appendix~\ref{app: rkhs regression}). 
The next lemma is a standard result by \cite{srinivas2009gaussian,abbasi2013online}. It provides 
a frequentist confidence interval of the form $[\hat{f}_t(x,c) \pm \beta_t \sigma_t(x,c)]$ that contains the true function values $f(x,c)$ with high probability. The exact definitions of $\hat{f}_t$ and $\sigma_t$ can be found in Appendix \ref{app: rkhs regression}; we just note here that $\hat{f}_t(x,c)$ and $\sigma_t(x,c)^2$ are the posterior mean and posterior variance functions of the corresponding Bayesian Gaussian process model \citep{rasmussen2006gaussian}. We denote the data kernel matrix by $(K_t)_{i,j=1,\dots,t} = k(x_i,c_i,x_j,c_j)$, and assume that $k(x,c,x',c') \leq 1$.

\begin{lemma} \label{lemma:confidence_bound}
%Let $\hat{f}_t$ be the least-squares estimator \eqref{eq: kernel ridge regression} for possibly adaptively collected data $\Dc_t$.
With probability at least $1-\delta$, for any $x\in \Xc$, $c \in \Cc$ at any time $t\geq 1$,
	\begin{align*}
|\hat{f}_t(x,c) - f(x,c)| \leq \beta_t \sigma_t(x,c)
	\end{align*}
	with $\beta_t = \sigma \sqrt{\log \det\big(\mathbf{1}_t + K_t\big) + 2\log\frac{1}{\delta}} + B$.
\end{lemma}

We explicitly define the upper and lower confidence bounds for every $x \in \Xc$ and $c \in \Cc$ as follows:
\begin{align*}	
	\ucb_t(x,c) & := \hat{f}_t(x,c) + \beta_t \sigma_{t}(x,c), \\
	\lcb_t(x,c) & := \hat{f}_t(x,c) - \beta_t \sigma_{t}(x,c).
\end{align*}
For a fixed $x$, we use $\ucb_{x}^t:= \ucb_t(x,\cdot)$ and $\lcb_{x}^t:= \lcb_t(x,\cdot)$ to refer to the corresponding vectors in $\R^{|\Cc|}$.

Finally, we introduce a sample complexity parameter, the \emph{maximum information gain}:
\begin{align}
\label{eq:mmi}
\gamma_T := \max_{ \lbrace (x_t,c_t) \rbrace^{T}_{t=1} }\log \det\big(\mathbf{1}_t + K_T\big) \text{ .}
\end{align}
% In the GP model, this corresponds to the maximum mutual information $\max_{x_1, \dots, x_T \in \Xc} \I(x_1,\dots, x_t; f)$. 
\looseness=-1 The information gain appears in the regret bounds for Bayesian optimization \citep{srinivas2009gaussian}. Analytical upper bounds are known for a range of kernels, e.g., for the RBF kernel, $\gamma_T \leq \Oc(\log(T)^{d+1})$ if $\Xc \times \Cc \subset \R^d$.

\paragraph{Maximum Mean Discrepancy (MMD).}

MMD is a kernel-based discrepancy measure between distributions (e.g.,~\cite{muandet2017kernel}). It has been used in various applications, including generative modeling~\citep{sutherland2016generative, binkowski2018demystifying}, DRO~\citep{staib2019kernel} and kernel sample tests~\citep{gretton2012kernel,chwialkowski2016kernel}. Let  $\Hc_M$ be an RKHS with corresponding kernel $k_M : \Cc \times \Cc \rightarrow \R$. For two distributions $P$ and $Q$ over $\Cc$, the \emph{maximum mean discrepancy} (MMD) is 
\begin{equation} \label{eq: mmd}
d(P,Q):= \sup_{g \in \Hc_M : \| g \|_{\Hc_M} \leq 1} \E_{c \sim P}[g(c)] - \E_{c \sim Q}[g(c)] \text{ .}   
\end{equation}
Note that the kernel $k_M$ over $\Cc$ that defines the MMD is different from the kernel $k$ over $\Xc \times \Cc$ that is used for regression. An equivalent way of writing $d(P,Q)$ is via \emph{kernel mean embeddings}~\cite[Section 3.5]{muandet2017kernel}. Specifically, any distribution $P$ over $\Cc$ can be embedded into $\Hc_M$ via the mean embedding $m_{P} = \E_{c\sim P}[k_M(c, \cdot)]$, which satisfies $\<m_{P}, k_M(c', \cdot)\> = \E_{c\sim P}[k_M(c', c)]$ for all $c' \in \Cc$. An equivalent expression for the MMD \eqref{eq: mmd} is
\begin{equation} 
d(P,Q) =  \| m_{P} - m_{Q} \|_{\Hc} \text{ .} \label{eq: mmd with kernel mean embeddings}
\end{equation}
More explicitly, for finite context set $\Cc$ and probability vectors $w_i = \P_P[c_i]$ and $w'_i = \P_Q[c_i]$, the kernel mean embeddings are $m_P = \sum_{i=1}^n w_i k_M(c_i,\cdot)$ and $m_Q = \sum_{i=1}^n w'_i k_M(c_i,\cdot)$, respectively. With the kernel matrix $(M)_{ij} := k_M(c_i,c_j)$, the MMD becomes
\begin{equation*}
d(P,Q)  = \sqrt{(w-w')^\T M (w-w')}  =:  \|w - w'\|_{M}\text{ .}
\end{equation*}

\begin{algorithm}[t]
	\begin{minipage}{\columnwidth}
		Initialize  $(K_x)_{i,j} = k(x,c_i,x,c_j), \Cc = \{c_1, \dots, c_n\}$ \\
		\textbf{For} step $t=1,2,\dots, T$:
		\begin{enumerate}
			\itemsep0em
			\item Learner obtains reference distribution $P_t$ with $w^t_i = \P[c = c_i]$, and margin $\epsilon_t$
			\item Define $(\ucb^t_x)_j := \hat{f}_t(x,c_j) + \beta_t \sigma_t(x,c_j)$ 
			\item Define $w_x^{\ucb_t} := \argmin_{w'} \< \ucb^t_x, w' \>$, s.t.~$\|w'\|_1 = 1,  0\leq w'_j \leq 1\; (\forall j \in [n]), \text{ and } \|w' - w^t\|_{M}\leq \epsilon_t$
			\item Choose action $x_t = \argmax_{x \in \Xc} \<w_x^{\ucb_t}, \ucb^t_x\>$
			\item Learner observes $c_t \sim P^*_t$ and $y_t = f(x_t,c_t) + \xi_t$. 
			\item Use $\lbrace x_t, c_t, y_t \rbrace$ to update $\hat{f}_{t+1}(\cdot,\cdot)$ and $\sigma_{t+1}(\cdot,\cdot)$.
		\end{enumerate}
		\caption{DRBO - General Setting}\label{alg: general drbo}
	\end{minipage}
\end{algorithm}

\section{Distributionally Robust Bayesian Optimization}
We now introduce a Bayesian optimization algorithm for our main objective \eqref{eq:robust_cumul_regret}. We will start with a \emph{general formulation} that allows for time-dependent reference distributions $P_t$ and margins $\epsilon_t$. We then continue with \emph{data-driven} DRO \citep{bertsimas2018data}, where we specialize the general setup and choose the empirical distribution $P_t = \frac{1}{t}\sum_{s=1}^t \delta_{c_s}$ as reference distribution. 
%A key property is that if $c_t \sim P^*$, then $d_{\text{MMD}}(P_t,P^*) \leq \Oc(\sqrt{\frac{1}{t}\log \frac{1}{\delta}})$. 
Hence, our algorithm chooses actions that are robust w.r.t.~the estimation error of the true context distribution. Finally, we motivate and discuss the \emph{simulator} setting, where the learner is allowed to choose the context $c_t$ and obtains the corresponding evaluation $y_t = f(x_t,c_t) + \xi_t$.

\subsection{General DRBO}
In our general DRBO formulation, the interaction protocol at time $t$ is specified by the following steps:
\begin{enumerate}
	\itemsep0em 
	\item The environment chooses a reference distribution $P_t$ and margin $\epsilon_t$. This defines the uncertainty set 
	\begin{equation}
	\Uc_t = \{ Q : d(Q,P_t) \leq \epsilon_t\} \text{ .}
	\end{equation}
	\item The learner observes $P_t$ and $\epsilon_t$, and chooses a robust action $x_t \in \Xc$.
	\item The environment chooses a sampling distribution $P_t^* \in \Uc_t$ and the context is realized as an independent sample $c_t \sim P_t^*$.
	\item The learner observes the reward $y_t = f(x_t,c_t) + \xi_t$ and $c_t \sim P_t^*$.
\end{enumerate} 

We make no further assumptions on how the environment chooses the sequences $P_t,P_t^*$ and $\epsilon_t$.
The DRBO algorithm for this setting is given in Algorithm~\ref{alg: general drbo}. Recall that $P_t$ is a distribution over the finite context set $\Cc$ with $n$ elements, and we use $w^t \in \R^{n}$ to denote a probability vector with entries $w^t_i = \P_{P_t}[c = c_i]$ for every $i \in [n]$. With this, the inner adversarial problem for a fixed action $x$ can be equivalently written as:
\begin{align} \label{eq:obj_in_terms_of_w}
\inf_{Q: d(P_t,Q) \leq \epsilon_t} \E_{c \sim Q}[f(x,c)]  = \min_{\substack{w' : \|w'\|_1 = 1,\\0\leq w'_j \leq 1\; \forall j \in [n],\\ \|w'-w^t\|_{M} \leq \epsilon_t}} \< w', f_x \>, 
\end{align}
where $f_x:=f(x,\cdot) \in \R^{n}$, and $M \in \R^{n \times n}$ with $(M)_{ij} := k_M(c_i,c_j)$. In particular the solution to \eqref{eq:obj_in_terms_of_w} is the worst-case distribution over $c$ for the objective $f$ if the learner chooses action $x$. Since the constraints are convex, the program \eqref{eq:obj_in_terms_of_w} can be solved efficiently by standard convex optimization solvers. 

Since the true function values $f_x$ are unknown to the learner, we can only obtain an approximate solution to \eqref{eq:obj_in_terms_of_w}. In our algorithm, we hence use an optimistic upper bound instead. Specifically, we substitute $\ucb^t_x$ for $f_x$ to compute the ``optimistic'' worst-case distribution for every action $x$. Finally, at time $t$, the learner chooses $x_t$ that maximizes the optimistic expected reward under the worst-case distribution.

The DRBO algorithm achieves the following regret bound.

\begin{theorem} \label{thm: regret general drbo}
	The robust regret $R_T$ of Algorithm \ref{alg: general drbo}, with $\beta_t= \sigma \sqrt{\log \det\big(\mathbf{1}_t + K_t\big) + 2\log\frac{2}{\delta}} + B$, is bounded with probability at least $1-\delta$ by
	\begin{equation*}
	R_T \leq 4 \beta_T\sqrt{T\big(\gamma_T+ 4 \log\left(\tfrac{12}{\delta}\right)\big)} + 2B' \sum_{t=1}^T \epsilon_t \text{ .}
	\end{equation*}
	Here, $\gamma_T$ is the maximum information gain defined in Eq.\ \eqref{eq:mmi}, $\|f\|_{\Hc} \leq B$ and $B' = \max_{x\in \Xc} \|f_x\|_{M^{-1}}$.
\end{theorem}

The complete proof is given in Appendix \ref{app: proof theorem data driven}, and we only sketch the main steps here. Denote by $w_t^*$ the probability vector of the true distribution at time $t$, and by $w^{f}_{x_t}$ the solution to \eqref{eq:obj_in_terms_of_w} at $x_t$. The idea is to bound the instantaneous regret at time $t$ by
\begin{align*}
r_t &= \inf_{Q: d(P_t,Q) \leq \epsilon_t} \E_Q[f(x^*,c)]- \inf_{Q: d(P_t,Q) \leq \epsilon_t} \E_Q[f(x_t,c)]\\
&\stackrel{(i)}{\leq} \< w_t^*, \ucb^t_{x_t}\>- \<w_{x_t}^{f}, f_{x_t} \>\\
&=  \< w_t^*, \ucb^t_{x_t}- f_{x_t}\> + \< w_t^* - w_{x_t}^{f}, f_{x_t}\>\\
&\stackrel{(ii)}{\leq} 2 \beta_t \< w_{t}^*, \sigma_t(x_t, \cdot) \> +  \|w_t^* - w_{x_t}^{f}\|_{M} \|f_{x_t}\|_{M^{-1}} \\
&\stackrel{(iii)}{\leq} 2 \beta_T \< w_{t}^*, \sigma_t(x_t, \cdot) \> + 2\epsilon_t B' \text{ .}
\end{align*}
For the first inequality (i), we used that $f_x \leq \ucb_x$, the definition of the UCB action and that $w_t^* \in \Uc_t$. In step (ii), we use Cauchy-Schwarz and the confidence bounds, and step (iii) follows since $w_{x_t}^f \in \Uc_t$. From here it remains to sum the instantaneous regret, where we rely on Lemma~3 in \citep{kirschner18heteroscedastic} to relate the expectation over the true sampling distribution $\< w_{t}^*, \sigma_t(x_t, \cdot) \>$ to the observed values  $\sigma_t(x_t, c_t)$. 

In the regret bound in Theorem~\ref{thm: regret general drbo}, the first term is the same as the standard regret bound for GP-UCB \citep{srinivas2009gaussian,abbasi2013online} and reflects the statistical convergence rate for estimating the RKHS function. The additional term $B'T\epsilon$ (for $\epsilon_t = \epsilon$) is specific to our setting. First, the complexity parameter $B' =  \max_{x\in \Xc} \|f_x\|_{M^{-1}}$ quantifies how much the distributional shift can increase the regret on the given objective $f$. A crude upper bound is $B' \leq B\sqrt{\lambda_{\max}(M^{-1}) |\Cc|}$, but in general $B'$ can be much smaller. The linear scaling $\Oc(\epsilon T)$ of the regret bound is arguably unsatisfying, but seems unavoidable without further assumptions. A problematic case is when the true distribution $P_t^*$ is supported on a single context, e.g., $P_t^* = \delta_{c_1}$, and the learner is not able to learn the function values at different contexts $c_i$ for $i > 1$. In this case, the learner can never infer the robust solution exactly from the data and consequently incurs constant regret of order $\epsilon_t$ per round. In practice, we do not expect that this severely affects the performance of our algorithm if the true distribution sufficiently \textit{covers} the context space. We leave a precise formulation of this intuition for future work. 

Instead, in the following sections we explore two different ways of controlling the additional regret that the learner incurs in the general DRBO setting. First, for the \emph{data-driven} setting, we will set the reference distribution to the empirical distribution of the observed context samples. In this case, the margin $\epsilon_t$ is the distance to the true sampling distribution, which for the MMD is of order $1/\sqrt{t}$ and results in $\sum_{t=1}^T \epsilon_t = \Oc(\sqrt{T})$. In the second variant, the learner is allowed to also choose $c_t$, which circumvents the estimation problem outlined above and avoids the linear regret term. \looseness=-1

\subsection{Data-Driven DRBO} \label{sec:dd_drbo}
\looseness=-1 In \emph{data-driven} DRBO, we assume there is a fixed but unknown distribution $P^{*}$ on $\Cc$. In each round, the learner first chooses an action $x_t \in \Xc$, and then observes a context sample $c_t \sim P^{*}$ together with the corresponding observation $y_t = f(x_t,c_t) + \xi_t$. At the beginning of round $t$, the learner compute the empirical distribution $\hat{P}_{t} = \frac{1}{t-1}\sum_{s=1}^{t-1} \delta_{c_s}$ using the observed contexts $\{c_1,\dots, c_{t-1} \}$. The objective is to choose a sequence of actions $x_t$, which is robust to the estimation error in $\hat{P}_t$. This corresponds to minimizing the robust regret \eqref{eq:robust_cumul_regret}, where we set $P_t = \hat{P}_t$ for every $t$. 

As the learner observes more context samples, she becomes more confident about the true unknown $P^*$. It is therefore reasonable to shrink the uncertainty set of distributions $\Uc_t = \{Q : d(Q,\hat{P}_t) \leq \epsilon_t\}$ over time. We make use of the following lemma. %The following lemma quantifies the convergence of the empirical distribution $\hat{P}_t$ to the true sampling distribution $P^{*}$ in MMD distance. 
\begin{lemma}[\cite{muandet2017kernel}, Theorem 3.4]\label{lemma: mmd convergence}
Assume $k(c_i,c_j) \leq 1$ for all $c_i, c_j \in \Cc$. Let $P^{*}$ be the true context distribution over $\Cc$, and let $\hat{P}_t = t^{-1} \sum_{s=1}^{t} \delta_{c_s}$ be the empirical sample distribution. Then, with probability at least $1 - \delta$,\looseness=-1
	\begin{equation*}
		d(P^{*}, \hat{P}_t) \leq \frac{1}{\sqrt{t}} \left(2+  \sqrt{2 \log(1 / \delta)}\right).
	\end{equation*}
\end{lemma}
Lemma~\ref{lemma: mmd convergence} shows how to set the margin $\epsilon_t$ such that, at time $t$, the true distribution is contained with high probability in the uncertainty set around the empirical distribution.
The interaction protocol at time $t$ is then: %summarized as follows:
\begin{enumerate}
	\itemsep0em 
	\item The learner computes the empirical distribution $\hat{P}_{t}$ and corresponding margin $\epsilon_{t}$ according to Lemma \ref{lemma: mmd convergence}, and defines the uncertainty set
	\begin{equation*}
	\Uc_t = \{ Q: d(Q,\hat{P}_{t}) \leq \epsilon_{t} \} \text{ .}
	\end{equation*} 
	\item The learner chooses a robust action $x_t$.
	\item The learner observes reward $y_t = f(x_t, c_t) + \xi_t$ and context sample $c_t \sim P^*$.
\end{enumerate}

We follow Algorithm~\ref{alg: general drbo}, and set the reference distribution and margin as outlined above. As a consequence of Theorem \ref{thm: regret general drbo} we obtain the following regret bound.
\begin{corollary}\label{cor: data driven}
The robust regret $R_T$ of Algorithm \ref{alg: general drbo}, with $\beta_t= \sigma \sqrt{\log \det\big(\mathbf{1}_t + K_t\big) + 2\log\frac{3}{\delta}} + B$ and $\epsilon_t = \frac{1}{\sqrt{t}} \left(2+  \sqrt{2 \log(\frac{6t^2}{\delta})}\right)$ is bounded in the data-driven scenario with probability at least $1-\delta$ by
	\begin{align} \nonumber
	R_T &\leq 2\beta_T \sqrt{T} \sqrt{\gamma_T (1 + \log(3/\delta)}  \\
		& \qquad + 4 B' \sqrt{T} \big(2 + \sqrt{2 \log\big(\tfrac{6T^2}{\delta}\big)}\big) \text{ ,}
	\end{align}
	where $\gamma_T$ is the maximum information gain as defined in \eqref{eq:mmi}, $\|f\|_{\Hc} \leq B$ and $B' = \max_{x\in \Xc} \|f_x\|_{M^{-1}}$.
\end{corollary}

The proof can be found in Appendix \ref{app: proof cor data driven}. We just note that we increased the value of $\epsilon_t$ such that Lemma~\ref{lemma: mmd convergence} holds simultaneously over all time steps. In the data-driven contextual setting \emph{without the robustness requirement}, several related approaches have been proposed \citep{lamprier2018profile,kirschner2019context}. These are based on computing a UCB score directly at the kernel mean embedding of the empirical distribution $\hat{P}_t$ . To account for the estimation error, an additional exploration bonus is added. We note that as $t \rightarrow \infty$ and $\hat{P}_t$ becomes an accurate estimation of $P^*$, both robust and non-robust approaches converge to the stochastic solution. The advantage of the \emph{robust} formulation is that we explicitly minimize the loss under the worst-case estimation error in the context distribution. As we demonstrante in our experiments (in Section~\ref{sec:experiments}), DRBO obtains significantly smaller regret when the robust and stochastic solutions are different.

\subsection{Simulator DRBO}

\begin{algorithm}[t]
	\begin{minipage}{\columnwidth}
		Initialize  $(K_x)_{i,j} = k(x,c_i,x,c_j), \Cc = \{c_1, \dots, c_n\}$ \\
		\textbf{For} step $t=1,2,\dots, T$:
		\begin{enumerate}
			\itemsep0em
			\item Obtain reference distribution $P_t$ with $w^t_i = \P[c = c_i]$, margin $\epsilon_t$
			\item Define $(\ucb^t_x)_j := \hat{f}_t(x,c_j) + \beta_t \sigma_t(x,c_j)$ 
			\item $w_x^{\ucb_t} := \argmin_{w'} \< \ucb^t_x, w' \>$, s.t.~$\|w'\|_1 = 1,  0\leq w_j \leq 1\; (\forall j \in [n]), \|w' - w\|_{M}\leq \epsilon_t$
			\item $x_t = \argmax_{x \in \Xc} \<w_x^{\ucb_t}, \ucb^t_x\>$
			\item $c_t = \argmax_{c \in \Cc} \sigma_t(x_t, c)$.
			\item Observe $y_t = f(x_t,c_t) + \xi_t$ from simulator
		\end{enumerate}
	\end{minipage}
	\caption{DRBO - Simulator Setting}\label{alg: drbo simulation seting}
\end{algorithm}

In our second variant of the general setup, the learner is allowed to choose $c_t$ in addition to $x_t$ and then obtains the observation $y_t = f(x_t,c_t) + \xi_t$.

One example of this setting, previously considered in the context of RO~\citep{bogunovic2018adversarially}, is when the learner tunes control parameters  with a simulator of the environment (e.g.\ for a building heating system). The simulator gives the learner the ability to evaluate the objective at any specific context $c_t$. The objective is to simultaneously (or only at the final time $T$) deploy a robust solution $x_T$ on the real system, where the covariate $c_t$ is uncontrolled. Again, the learner's objective is to be robust with respect to an uncertainty set of distributions on $c_t$ \emph{on the real environment} (e.g.\ for heating control, we want robustness on predicted weather conditions that effect the building's state). With this motivation in mind, we refer to this setup as \emph{simulator} DRBO. %\looseness=-1
Formally, the interaction protocol is:
\begin{enumerate}
	\itemsep0em 
	\item The environment provides a reference distribution $P_t$, margin $\epsilon_t$ and uncertainty set $\Uc_t$ as before.
	\item The learner chooses an action $x_t$ and a context $c_t \in \Cc$.
	\item The learner observes reward $y_t = f(x_t, c_t) + \xi_t$ from the simulator.
	\item The learner deploys a robust action $x_t$ on the real system (or possibly only at the final step $T$).
\end{enumerate}

%% begin experiments figure
\begin{figure}
	\includegraphics{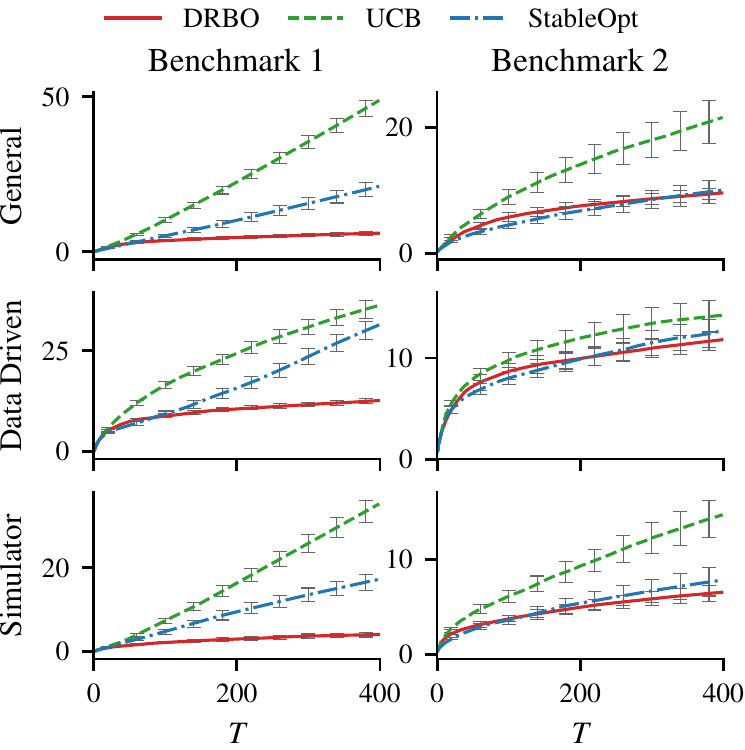}
	\caption{Results for two synthetic benchmarks, where the stochastic, worst-case robust and distributionally robust solution are all different \textit{(left)} or coincide \textit{(right)}.
		 All plots show robust regret, averaged over 50 independent runs and the error bars indicate the standard error.}\label{fig:synthetic}
\end{figure}

We provide Algorithm \ref{alg: drbo simulation seting} for this setting. As before, $x_t$ is an optimistic action under the worst-case distribution. In addition, the learner chooses $c_t = \argmax_{c \in \Cc} \sigma_t(x_t,c_t)$ as the context with the largest estimation uncertainty at $x_t$. We bound the robust regret in the next theorem.

\begin{theorem} \label{thm: regret simulator}
	In the simulator setting, Algorithm \ref{alg: drbo simulation seting}, with $\beta_t = \sigma \sqrt{\log \det\big(\mathbf{1}_t + K_t\big) + 2\log\frac{1}{\delta}} + B$, obtains bounded robust regret w.p.\ at least $1 - \delta$,
	\begin{equation*}
	R_T \leq 2\beta_T \sqrt{\gamma_T T}  \text{ .}
	\end{equation*}
\end{theorem}
We provide the proof of Theorem~\ref{thm: regret simulator} in Appendix \ref{app: proof regret simulator}.

Perhaps surprisingly, this rate is the same as for GP-UCB in the standard setting (a similar result was obtained for RO \citep{bogunovic2018adversarially}). This is because now the learner can estimate $\hat{f}_t$ globally at any input $(x_t,c_t) \in \Xc \times \Cc$, and the sample complexity to infer the robust solution only depends on the sample complexity of estimating $f$.\looseness=-1

In the simulator setting, the performance of the \emph{final solution} can be of significant interest if we aim to deploy the obtained parameter on the real system. To this end, we allow the final solution $\hat{x}_T$ to be different from the last evaluation $x_T$. The metric of interest is then the robust simple regret,
\begin{equation*}
r_T = \max_{x \in \Xc} \inf_{Q} \E_{c \sim Q}\; f(x, c) - \inf_Q \E_{c \sim Q} f(\hat{x}_T, c) \text{.}
\end{equation*}
To obtain a bound on the simple regret, we assume that the margin $\epsilon = \epsilon_t$ and the reference distribution $P = P_t$ are fixed. This is a natural requirement, which allows the learner to optimize the simple regret for the final solution $\hat{x}_T$ w.r.t.\ $P$ and $\epsilon$. We choose the final solution $\hat{x}_T := x_{\hat{t}}$ among the iterates $x_1, \dots, x_T$ from Algorithm \ref{alg: drbo simulation seting} with 
\begin{equation}\label{eq:solution}
\hat{t} := \argmax_{t=1, \dots, T} \min_{\substack{w' : \|w'\|_1 = 1,\\0\leq w'_j \leq 1\; \forall j \in [n],\\ \|w'-w^t\|_{M} \leq \epsilon}} \< w', \lcb^t_{x_t}\> \text{ .}
\end{equation}
The program computes the best robust solution among the iterates $\{x_1,\dots,x_T\}$ using the \emph{conservative} function values $\lcb^t_x$ of the corresponding time steps $t$. It is easy to maintain $\hat{x}_T$ iteratively by computing the conservative, worst-case payoff of the action $x_t$ and comparing to the previous solution $\hat{x}_{t-1}$.

\begin{corollary}[Simple Regret]\label{cor: simple regret}
	With probability at least $1 - \delta$, the solution $\hat{x}_T$ obtains simple regret
	\begin{align}
	r_T \leq 2\beta_T \sqrt{{\gamma_T}/{T}}. 
	\end{align}
\end{corollary}
	
This result is a consequence of the fact that the simple regret of $\hat{x}_t$ is upper bounded by the simple regret of each iterate $x_t$. The guarantee then follows from the proof of Theorem \ref{thm: regret simulator}. We provide the complete argument in Appendix \ref{app: proof simple regret simulator}.

\section{Experiments}
\label{sec:experiments}

We evaluate the proposed DRBO in the general, data-driven and simulator setting on two synthetic test functions, and on a real-world wind-power prediction task. In our experiments, we compare to StableOpt \citep{bogunovic2018adversarially} and a stochastic UCB variant \citep{srinivas2009gaussian,kirschner2019context}. \looseness=-1

\paragraph{Baselines} The first baseline is a stochastic variant of the UCB approach \citep{srinivas2009gaussian,kirschner2019context}, which chooses actions according to optimistic expected payoff w.r.t.\ the reference distribution,
\begin{equation*}
x_t^{\text{UCB}} = \argmax_{x \in \Xc} \E_{P_t}[\ucb_t(x,c)] \text{ .}
\end{equation*}
Our second baseline is StableOpt \citep{bogunovic2018adversarially}, an approach for worst-case robust optimization. It chooses actions according to 
\begin{equation*}
x_t^{\text{STABLE}} = \argmax_{x \in \Xc} \min_{c \in \Delta_t} \ucb_t(x,c) \text{ ,}
\end{equation*}
for a robustness set of possible context values $\Delta_t \subset \Cc$. There is no canonical way of choosing $\Delta_t$ in our setting, and we use $\Delta_t = \{ c \in \Cc : \|c - \E_{c' \sim P_t}[c']\|_2 \leq \epsilon_t \}$. With the decreasing margin and the discretization of the context domain, it can happen that $\Delta_t$ is an empty set. In this case we explicitly set $\Delta_t = \{\argmin_{c \in \Cc} \|c - \E_{c' \sim P_t}[c']\|_2 \}$.

UCB and StableOpt optimize for the \emph{stochastic and worst-case robust solutions} respectively, and therefore can exhibit \emph{linear} regret for the robust regret (unless $\epsilon_t \rightarrow 0$ as in the data-driven setting). For all approaches we use the same RKHS hyper-parameters. In particularly we set $\beta_t=2$, which is a common practice to improve performance over the (conservative) theoretical values.

\paragraph{Benchmarks}

Our \emph{first synthetic benchmark} is the function illustrated in the introduction. The reference distribution is $P_t = \Nc(0.5, 0.05)$ and the true sampling distribution is $P^* = \Nc(0.45,0.1)$. For simplicity, we set the margin to the exact MMD distance $\epsilon_t := d(P_t, P^*)$. On this function, the stochastic, worst-case robust and distributionally robust solution all differ, which leads to linear robust regret for UCB and StableOpt.  The \emph{second synthetic benchmark} is chosen such that stochastic, worst-case and distributionally robust solutions coincide, with the same choice of $P_t, P^*$ and $\epsilon_t$ as before. See Appendix \ref{app: experiments}, Fig.\ \ref{fig: benchmark 2 illustration} for a contour plot. Fig.\ \ref{fig:synthetic} illustrates the results.

\begin{figure}[t]
	\centering
	\includegraphics{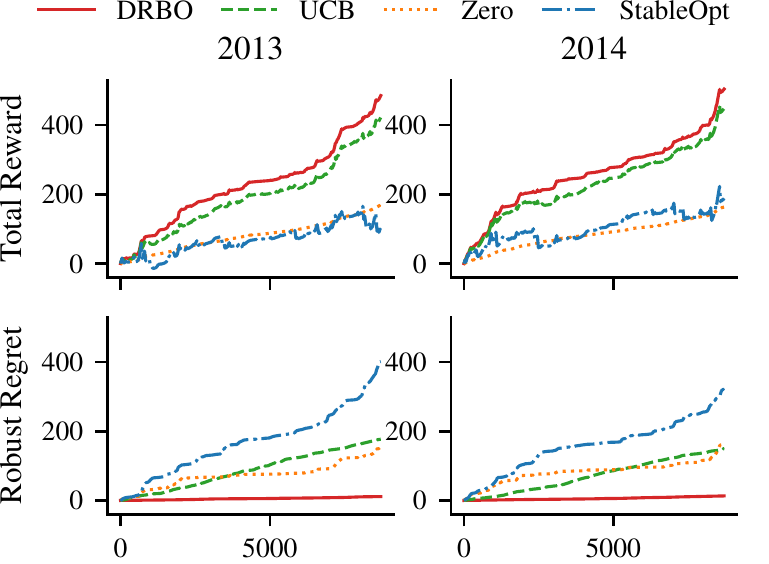}
	\caption{\emph{Wind power prediction}. We show cumulative revenue (top) and robust regret (bottom). StableOpt is too conservative to perform well on either objective. UCB does not account for the distributional shift on the sliding window. By definition, DRBO chooses the robust solution most of the time, therefore achieves (almost) zero robust regret.}
	\label{fig: wind power}
\end{figure}

\looseness=-1 Further, we evaluate the methods on real-world wind power data \citep{winddata2019}. Wind power forecasting is an important task \citep{wang2011review} as power sources that can be effectively scheduled are valuable on the global energy market. In our problem setup, we take hourly recorded wind power data from 2013/14 and use a 48h sliding window to compute an empirical reference distribution for each time step. The decision variable $x$ is the amount of energy that is guaranteed to be delivered in the next hour after the end of the window. The contextual variable  $c$ is the actual power generation which we take from the data set. We choose the reward (revenue) function: 
\[f(x, c) = 0.1\max(c - x, 0) + \min(x, c) - 5\max(x - c, 0)\,.\]
There is a $0.1$ reward/energy that was not committed ahead of time, $1$ reward/energy for committed energy and $-5$ penalty for committed energy that is not delivered (if the wind generation was too low). For each time step, we use the simulator scenario to compute the robust/stochastic/stable solution; and evaluate the performance on the data set.
In Figure \ref{fig: wind power}, we report the cumulative revenue of the different solutions deployed at each time step; this corresponds to the total revenue obtained during the year. The additional baseline is a ``zero commitment strategy'' ($x_t=0$). The figure also shows cumulative robust regret. Clearly, the stochastic solution is different from the robust one, hence UCB obtains linear robust regret. In fact, in this case if the DRO objective is solved exactly for each step, the DRBO method would obtain zero robust regret (we compute the solution according to \eqref{eq:solution} after $T=100$ steps, therefore an optimization error may remain).

\section{Conclusion}
In this work, we introduced and studied distributionally robust Bayesian optimization, where the goal is to be robust against the worst-case contextual distribution among a specified uncertainty set of distributions. Specifically, we focused on uncertainty sets determined by the MMD distance. For a few settings of interest that differ in how the contextual parameter is realized, we provided the first DRBO algorithms with theoretical guarantees. In the experimental study, we demonstrated improvements in terms of robust expected regret over stochastic and worst-case BO baselines. 

\looseness=-1 Our algorithms rely on solving the inner adversary problem, which, in our case, is a linear program with convex constraints. This program can be solved efficiently but is of size $|\Cc|$, which currently limits the method to relatively small context sets. The formulation and the theory continue to hold for large or continuous context sets, but finding a tractable algorithmic approximation is an interesting direction for future work. Finally, while the considered kernel-based MMD distance fits well with the kernel-based regularity assumptions used in BO, an interesting direction is to extend the ideas to other uncertainty sets used in machine learning, such as the ones defined by $\phi$-divergences and Wasserstein distance. In fact, our approach is still applicable in the case of other divergences, as long as the uncertainty set of distributions is convex and the inner problem can be solved efficiently.

\section*{Acknowledgement}
This project has received funding from the
 European Research Council (ERC) under the European Union’s Horizon 2020 research, innovation
 programme grant agreement No 815943, and NSF CAREER award 1553284.
IB is supported by ETH Z\"urich Postdoctoral Fellowship 19-2 FEL-47.

\bibliography{mybib}
\bibliographystyle{apalike}

\onecolumn
\newpage
\appendix

\section{RKHS Regression}\label{app: rkhs regression}
Recall that at step $t$, we have data $\Dc_t = \{(x_1,c_1,y_1), \dots, (x_t,c_t,y_t)\}$. The kernel ridge regression estimate is defined by,
\begin{equation}
\hat{f}_t = \argmin_{g \in \Hc} \sum_{i=1}^t (g(x_i,c_i) - y_i)^2 + \|g\|_{\Hc}^2 \text{ .}
\end{equation}
Denote by $\yb_t = [y_1, \dots, y_t]^\T$ the vector of observations, $(K_t)_{i,j=1,\dots,t} = k(x_i,c_i,x_j,c_j)$ the data kernel matrix, and $k_t(x,c) = [k(x,c,c_1,x_1), \dots, k(x,c,x_t,c_t)]^\T$ the data kernel features. We then have
\begin{align}
\hat{f}_t(x,c) &= k_t(x,c)^\T(K_t + \mathbf{1}_t)^{-1}\yb_t \text{ .}
\end{align}
We further have the posterior variance $\sigma_t(x,c)^2$ that determines the width of the confidence intervals,
\begin{align}
\sigma_t(x,c)^2 &= k(x,c,x,c) - k_t(x,c)^T(K_t + \mathbf{1}_t)^{-1} k_t(x,c) \text{ .}
\end{align}

\section{Proofs}
\subsection{Proof of Theorem \ref{thm: regret general drbo}}\label{app: proof theorem data driven}
The robust cumulative regret is
\begin{equation}
	R_T = \sum_{t=1}^T \max_{x \in \Xc} \inf_{Q : d(Q,P_t) \leq \epsilon_t} \E_Q[f(x,c)] - \inf_{Q : d(Q,P_t) \leq \epsilon_t} \E_Q[f(x_t,c)].
\end{equation}
For the proof, we first bound the \emph{instantaneous robust regret},
\begin{equation}
	r_t = \inf_{Q: d(P_t,Q) \leq \epsilon_t} \E_Q[f(x_t^*,c)] - \inf_{Q: d(P_t,Q) \leq \epsilon_t} \E_Q[f(x_t,c)] \text{ ,} %\leq 2\beta_t \sigma_t(x_t,c_t)\text{ .}
\end{equation}
where we denote $x_t^* =  \argmax_{x \in \Xc} \inf_{Q: d(P_t,Q) \leq \epsilon_t} \E_Q[f(x,c)]$ the true robust solution at time $t$.  We recall the following notation, $f_x = f(x,\cdot)$, $\lcb^{t}_x = \lcb_t(x,\cdot)$ and $\ucb^{t}_x = \ucb_t(x,\cdot)$ are vectors in $\R^n$, and $(M)_{i,j} = k(c_i, c_j)$. Further, $w_i = \P_P[c = c_i]$ is a probability vector in $\R^n$, where $n$ is used to denote the size of the contextual set, i.e., $n = |\Cc|$. With this, note that
\begin{align} \label{eq:obj_in_terms_of_w_1}
\inf_{Q: d(P_t,Q) \leq \epsilon_t} \E[f(x,c)]  = \inf_{\substack{w' : \|w'\|_1 = 1,\\0\leq w'_j \leq 1\; \forall j \in [n],\\ \|w'-w_t\|_{M} \leq \epsilon_t}} \< w', f_x \> 
\end{align}
The solution to this linear program is the worst case distribution over $c$ if we choose action $x$. Define worst-case distributions $w_x^f$, $w_x^{\lcb_t}$ and $w_x^{\ucb_t}$ for exact, optimistic and pessimistic function values (the dependence on $t$ is implicit),
\begin{align} \label{eq:w_ucb_and_lcb}
w_x^{f} = \argmin_{\substack{w' : \|w'\|_1 = 1,\\0\leq w'_j \leq 1 \; \forall j \in [n],\\ \|w' - w_t\|_{M}\leq \epsilon_t}} \< w', f_x \>,
\qquad
 w_x^{\lcb_t} = \argmin_{\substack{w' : \|w'\|_1 = 1,\\0\leq w'_j \leq 1 \; \forall j \in [n],\\ \|w' - w_t\|_{M} \leq \epsilon_t}} \< w', \lcb^t_x \>, 
\qquad 
w_x^{\ucb_t} = \argmin_{\substack{w' : \|w'\|_1 = 1,\\0\leq w'_j \leq 1 \; \forall j \in [n],\\ \|w' - w_t\|_{M}\leq \epsilon_t}} \< w', \ucb_x^t \> \text{ .}
\end{align} 
By combining \eqref{eq:obj_in_terms_of_w} and \eqref{eq:w_ucb_and_lcb}, we can upper and lower bound the objective as follows:
\begin{align} \label{eq:upper_and_lower_objective_bound}
\< w_x^{\lcb_t}, \lcb^t_x \> \leq \inf_{\substack{w' : \|w'\|_1 = 1,\\0\leq w'_j \leq 1\; \forall j \in [n],\\ \|w'-w_t\|_{M} \leq \epsilon_t}} \< w', f_x \> \leq \<  w_x^{\ucb_t}, \ucb^t_x \>.
\end{align}
Recall that Algorithm \ref{alg: general drbo} takes actions $x_t = \argmax_{x \in \Xc} \<w_{x}^{\ucb_t}, \ucb^t_x\>$, and note that $\|w_{x_t}^{\lcb_t} - w_t^{*} \|_{M} \leq \epsilon_t$ where $(w_t^{*})_i = \P_{P^{*}_t}[c=c_i]$ is the probability vector from the true sampling distribution at time $t$. 
%
%We use the notation $\min(a,b) = a \wedge b$ and $\max(a,b) = a \vee b$. First note that by bound on the RHKS norm we have $|f(x,c)| \leq \|f\| \leq B$. Hence we can always tighten the confidence bounds with
%\begin{align}
%\lcb(x,c) \vee -B \leq f(x,c) \leq \ucb(x,c) \wedge B \text{ .}
%\end{align}
%
For any $x\in \Xc$, we proceed to bound the instantaneous regret,

\begin{align}
r_t &= \inf_{Q: d(P,Q) \leq \epsilon} \E_Q[f(x_t^*,c)]- \inf_{Q: d(P,Q) \leq \epsilon} \E_Q[f(x_t,c)]\\
&\stackrel{(i)}{\leq} \< w_{x_t^*}^{\ucb_t}, \ucb^t_{x_t^*} \>- \<w_{x_t}^{f}, f_{x_t} \>\\
&\stackrel{(ii)}{\leq} \<w_{x_t}^{\ucb_t}, \ucb^t_{x_t} \>- \<w_{x_t}^{f}, f_{x_t} \>\\
&\stackrel{(iii)}{\leq} \< w_t^*, \ucb^t_{x_t} \>- \<w_{x_t}^{f}, f_{x_t} \>\\
&= \< w_t^*, f_{x_t}\> +  \< w_t^*, \ucb^t_{x_t}- f_{x_t}\>- \<w_{x_t}^{f}, f_{x_t} \>\\
&=  \< w_t^*, \ucb^t_{x_t}- f_{x_t}\> + \< w_t^* - w_{x_t}^{f}, f_{x_t}\>\\
&\stackrel{(iv)}{\leq} 2 \beta_t \< w_{t}^*, \sigma_t(x_t, \cdot) \> +  \|w_t^* - w_{x_t}^{f}\|_{M} \|f_{x_t}\|_{M^{-1}} \\
&\stackrel{(v)}{\leq} 2 \beta_t \< w_{t}^*, \sigma_t(x_t, \cdot) \>  + 2\epsilon_t \|f_{x_t}\|_{M^{-1}} \\
&\leq 2 \beta_T \< w_{t}^*, \sigma_t(x_t, \cdot) \> + 2\epsilon_t B' \text{ .}
\end{align}
Here, (i) follows from the definition of the upper bound \eqref{eq:upper_and_lower_objective_bound}, (ii) is by the choice of $x_t$, (iii) follows from the fact that $w_{x_t}^{\ucb_t}$ is a minimizer, (iv) uses again the confidence bounds and (v) follows from $d_{x_t}(P^*, Q) \leq \epsilon_t$. Finally, the following holds for the instantaneous regret: $r_t = \max_{x \in \Xc} r_t(x) \leq 2 \beta_T \< w_{t}^*, \sigma_t(x_t, \cdot) \> + 2 \epsilon_t B'$.

We now continue to bound the cumulative regret $R_T = \sum_{t=1}^T r_t$. To this end, we first apply the Cauchy-Schwarz inequality as in the standard proof, and then Jensen's inequality to find,
\begin{align}
R_T &\leq 2\beta_T \sum_{t=1}^T \< w_{t}^*, \sigma_t(x_t, \cdot) \> + 2B' \sum_{t=1}^T \epsilon_t\\ 
&\leq 2 \beta_T\sqrt{T \sum_{t=1}^T  \< w_{t}^*, \sigma_t(x_t, \cdot) \>^2} +2B' \sum_{t=1}^T \epsilon_t\\
&\leq 2\beta_T\sqrt{T \sum_{t=1}^T  \< w_{t}^*, \sigma_t(x_t, \cdot)^2 \>}+2B' \sum_{t=1}^T \epsilon_t \text{ .}
\end{align}

To complete the proof, we need to relate $\sum_{t=1}^T\< w_{t}^*, \sigma_t(x_t, \cdot) \>^2$ to the observed posterior variance $\sum_{t=1}^T \sigma_t(x_t, c_t)^2$. For this, we apply Lemma \ref{lemma: concentration of conditinal means} below. Note that $\sigma_t(x_t,c_t)^2 \leq 1$ by our assumption that $k(x,c,x',c')\leq 1$. Hence, w.p.\ at least $1-\delta$,
\begin{align}
\sum_{t=1}^T  \< w_{t}^*, \sigma_t(x_t, \cdot)^2 \> \leq 2\sum_{t=1}^T   \sigma_t(x_t, c_t)^2 + 8 \log\left(\frac{6}{\delta }\right)
\end{align}
Finally, using that $x \leq 2\alpha \log(1+x)$ for all $x \in [0, \alpha]$,
\begin{align}
 \sum_{t=1}^T \sigma_t(x_t, c_t)^2 \leq \sum_{t=1}^T  2 \log\left(1 + \sigma_t(x_t, c_t)^2\right)  \leq 2 \gamma_T \text{ .}
\end{align}
The last inequality follows from Lemma 3 in \cite{chowdhury17kernelized}. The final regret bound is therefore,
\begin{align}
R_T &\leq 4\beta_T \sqrt{T\big(\gamma_T+ 4 \log\left(\tfrac{6}{\delta}\right)\big)} +2 B' \sum_{t=1}^T \epsilon_t \text{,}
\end{align}
A union bound over the events such that both Lemma \ref{lemma:confidence_bound} and Lemma \ref{lemma: concentration of conditinal means} hold, yields probability $\geq 1-2\delta$ for the complete statement and completes the proof.

\begin{lemma}[Concentration of conditional mean, Lemma 3 in \cite{kirschner18heteroscedastic}]\label{lemma: concentration of conditinal means} Let $S_t \geq 0$ be non-negative stochastic process adapted to a filtration $\{\Fc_t\}$, and define $m_t = \E[S_t|\Fc_{t-1}]$. Further assume that $S_t \leq B$ for $B \geq 1$.
	% and let $(l_t)_{t\geq 1}$ be any fixed, positive sequence. 
Then, for any $T \geq 1$, with probability at least $1-\delta$ it holds that,
	\begin{align*}
	\sum_{t=1}^T m_t &\leq 2\sum_{t=1}^T S_t + 4 B \log\frac{1}{\delta} + 8  B\log(4 B) + 1\\
	& \leq 2\sum_{t=1}^T S_t + 8 B \log\frac{6B}{\delta}
	\end{align*}
\end{lemma}

\subsection{Proof of Corollary \ref{cor: data driven}} \label{app: proof cor data driven}

Note that with $\delta_t = \frac{2\delta}{\pi^2t^2}$, the result from Lemma \ref{lemma: mmd convergence} holds with probability at least $1-\delta/3$ by the union bound over all steps $t=1,2,3\dots$, i.e.
\begin{align}
d(\hat{P}_t, P^*) \leq \frac{1}{\sqrt{t}} \left(2+  \sqrt{2 \log\frac{\pi^2 t^2}{2\delta}}\right) = \epsilon_t
\end{align}
Therefore, with probability at least $1-\delta/3$, $P^* \in \Uc_t$. The results follows from Theorem \ref{thm: regret general drbo} and another application of the union bound. Finally we use that $\sum_{t=1}^T t^{-1/2} \leq 2\sqrt{T}$ to complete the proof of the corollary.

\subsection{Proof of Theorem \ref{thm: regret simulator}} \label{app: proof regret simulator}

Recall that Algorithm \ref{alg: drbo simulation seting} takes the actions $x_t = \argmax_{x \in \Xc} \<w_{x}^{\ucb_t}, \ucb^t_x\>$ and $c_t = \argmax_{c \in \Cc} \sigma_t(x_t, c)$. We begin to bound $r_t$ similar as in the proof of the general regret bound.
\begin{align}
	r_t(x) &= \inf_{Q: d(P_t,Q) \leq \epsilon_t} \E_Q[f(x,c)]- \inf_{Q: d(P_t,Q) \leq \epsilon_t} \E_Q[f(x_t,c)]\\
&\stackrel{(i)}{\leq} \< w_x^{\ucb_t}, \ucb^t_x \>- \<w_{x_t}^{\lcb_t}, \lcb^t_{x_t} \>\\
&\stackrel{(ii)}{\leq} \<w_{x_t}^{\ucb_t}, \ucb^t_{x_t} \>- \< w_{x_t}^{\lcb_t}, \lcb^t_{x_t} \>\\
&\stackrel{(iii)}{\leq} \<w_{x_t}^{\lcb_t}, \ucb^t_{x_t} \>- \< w_{x_t}^{\lcb_t}, \lcb^t_{x_t} \>\\
&\stackrel{(iv)}{\leq} 2\beta_t \sigma_t(x_t,c_t)
\end{align}
Here, as before, (i) replaces the function values by over/under-estimated values of the upper/lower confidence bounds, (ii) uses the choice of the UCB action and (iii) uses that $w_{x_t}^{\ucb_t}$ is a minimizer of $\<w_{x_t}^{\ucb_t}, \ucb^t_{x_t} \>$. The last inequality (iv) uses that $c_t$ maximizes $\sigma_t(x_t,c_t)$ as well as that $w_{x_t}^{\lcb_t}$ is a probability vector. With that, the cumulative regret bound follows via the standard argument.

\subsection{Proof of Corollary \ref{cor: simple regret}} \label{app: proof simple regret simulator}
To bound the simple regret% $r_T = \max_{x \in \Xc} \inf_Q \E_{Q}f(x,c) - \inf_Q \E_{Q}f(\hat{x}_T,c)$
, recall that $\hat{t} = \argmax_{t=1,\dots,T} \inf_Q \E_{Q} [\lcb_t(x_t,c)]$ and $\hat{x}_T = x_{\hat{t}}$. For any $t=1,\dots,T$ we have that,
\begin{align}
r_T &= \max_{x \in \Xc} \inf_{Q: d(Q,P) \leq \epsilon} \E_{Q}f(x,c) - \inf_{Q: d(Q,P) \leq \epsilon} \E_{Q}f(\hat{x}_T,c)\\
&\stackrel{(i)}{\leq}\max_{x \in \Xc} \< w_x^{\ucb_t}, \ucb^t_x \>- \<w_{\hat{x}_T}^{\lcb_{\hat{t}}}, \lcb^{\hat{t}}_{\hat{x}_T} \>\\
&\stackrel{(ii)}{=} \max_{x \in \Xc}\< w_x^{\ucb_t}, \ucb^t_x \>-  \max_{s=1,\dots,T}\<w_{x_s}^{\lcb_{s}}, \lcb^{s}_{x_s} \>\\
&\stackrel{(iii)}{\leq} \<w_{x_t}^{\ucb_t}, \ucb^t_{x_t} \>- \< w_{x_t}^{\lcb_t}, \lcb^t_{x_t} \>\\
&\stackrel{(iv)}{\leq} \<w_{x_t}^{\lcb_t}, \ucb^t_{x_t} \>- \< w_{x_t}^{\lcb_t}, \lcb^t_{x_t} \>\\
%&\stackrel{(iv)}{\leq} 2\beta_T \sigma_T(x_T,c_t)\\
&\stackrel{(v)}{\leq} 2\beta_t \sigma_t(x_t,c_t)
\end{align}
Here, (i) bounds the function values by the upper and lower confidence bound respectively, (ii) uses the definition of $\hat{x}_T$, (iii) uses the definition of the UCB action and drops the maximum and finally, (iv,v) as before uses the choices of $x_t$, $c_t$, $w_{x_t}^{\ucb_t}$ and that $w_{x_t}^{\lcb_t}$ is a probability vector. With this we are able to leverage the cumulative regret bound, and find
\begin{align}
r_T \leq \frac{1}{T} \sum_{t=1}^T 2\beta_t \sigma_t(x_t,c_t) \leq \frac{R_T}{T} \text{ .}
\end{align}

\section{Details on the Experiments} \label{app: experiments}

\begin{figure}[h]
	\begin{subfigure}[t]{0.45\columnwidth}
		\includegraphics{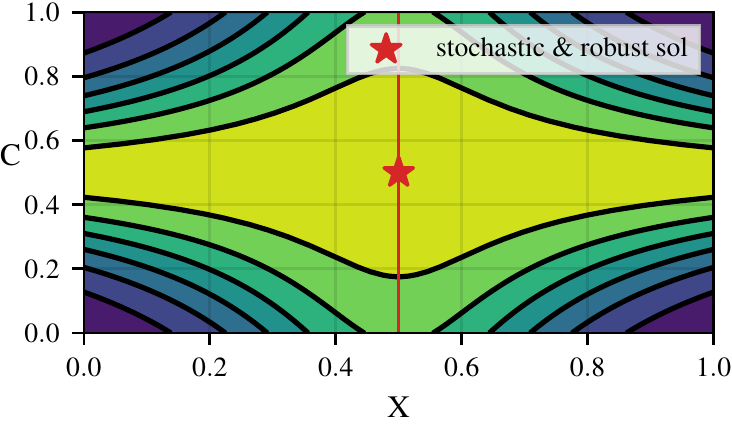}
		\caption{Second synthetic benchmark. Here stochastic and robust solutions coincide at the marked design.}\label{fig: benchmark 2 illustration}
	\end{subfigure}\hspace{20px}

	\label{fig: benchmark illustrations}
\end{figure}

% DO NOT INCLUDE THIS FOR SUBMISSION
%\input{parts/appendix_extra.tex}

\end{document}